\documentclass[10pt,twocolumn,letterpaper]{article}

\usepackage[preprint]{iccv}      

\definecolor{iccvblue}{rgb}{0.21,0.49,0.74}
\usepackage[pagebackref,breaklinks,colorlinks,allcolors=iccvblue]{hyperref}
\usepackage{algorithm}
\usepackage{algorithmic}

\usepackage{tabularx}

\newcommand{\tabincell}[2]{\begin{tabular}{@{}#1@{}}#2\end{tabular}}
\def\paperID{11402} 
\def\confName{ICCV}
\def\confYear{2025}

\title{VideoAgent2: Enhancing the LLM-Based Agent System for Long-Form Video Understanding by Uncertainty-Aware CoT}

\author{
Zhuo Zhi\textsuperscript{1}, Qiangqiang Wu\textsuperscript{2}, Minghe shen\textsuperscript{1}, Wenbo Li\textsuperscript{3}, Yinchuan Li\textsuperscript{3}, Kun Shao\textsuperscript{3},  Kaiwen Zhou \textsuperscript{3}\\
\\[1ex]
\textsuperscript{1}University College London\\
\textsuperscript{2}City University of Hong Kong\\
\textsuperscript{3}Huawei Noah’s Ark Lab\\
}
\begin{document}
\maketitle
\makeatletter
\renewcommand\@makefnmark{}
\makeatother
\footnotetext{Preprint.}
\begin{abstract}
Long video understanding has emerged as an increasingly important yet challenging task in computer vision. Agent-based approaches are gaining popularity for processing long videos, as they can handle extended sequences and integrate various tools to capture fine-grained information. However, existing methods still face several challenges: (1) they often rely solely on the reasoning ability of large language models (LLMs) without dedicated mechanisms to enhance reasoning in long video scenarios; and (2) they remain vulnerable to errors or noise from external tools.
To address these issues, we propose a specialized chain-of-thought (CoT) process tailored for long video analysis. Our proposed CoT with plan-adjust mode enables the LLM to incrementally plan and adapt its information-gathering strategy. We further incorporate heuristic uncertainty estimation of both the LLM and external tools to guide the CoT process. This allows the LLM to assess the reliability of newly collected information, refine its collection strategy, and make more robust decisions when synthesizing final answers. Empirical experiments show that our uncertainty-aware CoT effectively mitigates noise from external tools, leading to more reliable outputs. We implement our approach in a system called VideoAgent2, which also includes additional modules such as general context acquisition and specialized tool design. Evaluation on three dedicated long video benchmarks (and their subsets) demonstrates that VideoAgent2 outperforms the previous state-of-the-art agent-based method, VideoAgent, by an average of 13.1$\%$ and achieves leading performance among all zero-shot approaches.
\end{abstract}
\vspace{-6mm}    
\section{Introduction}
\label{sec:intro}
\begin{figure}
    \centering
    \includegraphics[width=1\linewidth]{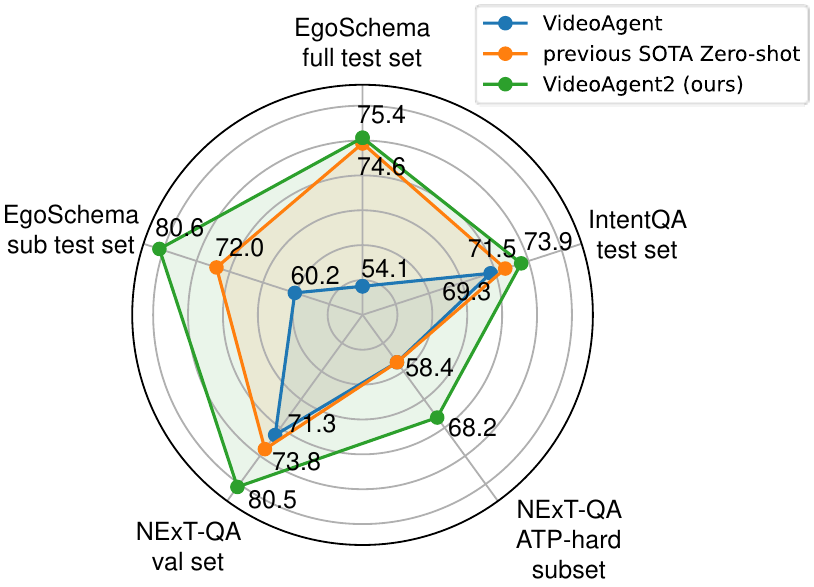}
    \caption{Performance of previous SOTA agent-based method VideoAgent \cite{wang2024videoagent}, previous SOTA Zero-shot method \cite{zhang2024hcqa, wang2023lifelongmemory, zhang2023simple, wang2024videoagent, ayyubi2025enter} and our proposed VideoAgent2 on all evaluation datasets. The metric is accuracy.}
    \label{fig:radar_map}
\end{figure}
Long video understanding has become a critical area in computer vision due to the increasing prevalence of lengthy video content across various domains, including video entertainment platforms, surveillance systems, and autonomous driving \cite{dai2023effect}. Unlike short video clips, long videos pose unique challenges, such as the need to model temporal dependencies over extended periods while maintaining affordable computational cost.

Although multimodal large language models (MLLMs) have shown initial success with short video clips \cite{li2024mvbench, wang2024internvideo2, jin2024video, damonlpsg2024videollama2, VoT24Hao}, they still face significant difficulties in processing long videos. Two primary challenges stand out. First, computational resources heavily limit the ability to process long videos in the same way as short clips. Most MLLMs are based on transformer architectures, which struggle to handle very long context inputs \cite{beltagy2020longformer}.

Some approaches have attempted to address this by compressing frame/context information \cite{he2024ma, li2023llama, song2024moviechat}. However, these methods risk over-compression, leading to information loss. Another challenge is that an MLLM's perception of fine-grained information is constrained by its encoder and training data. Capturing spatio-temporal information at various levels of granularity remains difficult. For instance, in question-answering tasks, different user queries require information at varying levels of granularity, which can differ significantly. Several solutions have been proposed, such as fine-grained instruction fine-tuning \cite{huang2024vtimellm}, chain-of-thought (CoT) reasoning \cite{fei2024video}, and specialized spatial and temporal information processing modules \cite{damonlpsg2024videollama2}. However, these methods often struggle with training complexity and slow reasoning speeds.

Recently, agent-based approaches have provided new insights for long video understanding \cite{zhang2024mm, wang2024videoagent, fan2024videoagent, wang2024vila, yu2024self, wang2024videotree}. Most of these methods utilize an LLM as an agent to analyze and reason about information retrieved from pre-trained image/video processing tools, thereby maximizing the strengths of LLMs in textual reasoning while maintaining computational efficiency. For instance, the state-of-the-art (SOTA) agent-based method VideoAgent \cite{wang2024videoagent} mimics the human video-watching process by first obtaining general context information, then retrieving relevant frames based on the user’s question, and iteratively repeating this process until sufficient information is gathered for the LLM to answer the question. Agent-based approaches avoid directly processing the entire video, thus sidestepping the challenges associated with long video processing. Moreover, the combination of various tools enables the acquisition of information at arbitrary levels of granularity.

However, some issues persist with the agent-based approach: (1) These methods rely heavily on the LLM's reasoning ability but focus primarily on system architecture design, neglecting how to improve LLM reasoning in long video scenarios \cite{fan2024videoagent, wang2024videoagent}. (2) The tools invoked by the LLM may introduce noise and 
hallucination, which can negatively impact LLM's reasoning and decision-making \cite{zhi2025seeing}.

To address these issues, we propose \textbf{VideoAgent2}, a system designed to enhance long video understanding with better accuracy and robustness. Specifically, we introduce a specialized CoT process for long-video agent systems that helps improve LLM's reasoning and decision-making capabilities. This process mimics human video understanding, using a plan-adjust mode to incrementally acquire information from coarse to fine-grained details. In addition, we incorporate uncertainty into the CoT process by leveraging heuristic uncertainty quantification from the LLM and the tools to guide reasoning. This approach mitigates noise and hallucination in both the LLM and tools, allowing the LLM to refine its information retrieval strategy and make
more reliable decisions when synthesizing final answers. Notably, this method does not require additional parameters or inference steps, making it more efficient than existing CoT methods \cite{zhang2406rest, setlur2024rewarding}. Finally, we propose a new pipeline for implementing VideoAgent2, incorporating important components such as general context information acquisition and specialized tool design.
Our main contributions are summarized as follows:
\begin{itemize}
\item We design a specialized CoT process based on the plan-adjust mode to enhance LLM reasoning and decision-making in long video understanding.
\item  We introduce uncertainty-guided CoT reasoning, which mitigates noise and hallucination in the system while requiring no additional parameters.
 \item We propose the VideoAgent2 pipeline, which integrates innovative designs such as general context acquisition and specialized tools. Our extensive evaluation on well-established long-form video understanding benchmarks, including Ego-Schema \cite{mangalam2023egoschema}, NExT-QA \cite{xiao2021next} and IntentQA \cite{li2023intentqa}, demonstrates VideoAgent2’s superior performance compared to existing methods. The results are illustrated in Fig. \ref{fig:radar_map}.
\end{itemize}
The paper is organized as follows. Section 2 gives an overview of related research. Section 3  introduces the proposed method. Section 4 shows the experiment results and the analysis. Finally, Section 5 summarizes the paper, its limitations, and future work.
\section{Related work}
\subsection{MLLM for long-form video understanding}
Significant efforts have been made to extend multimodal large language models (MLLMs) to video understanding, addressing two main challenges: the computational burden of long videos and the need to capture fine-grained spatio-temporal information. In LLaMA-VID, each video frame is represented as a content token and a context token to reduce the number of embedded tokens \cite{li2025llama}. MA-LLM employs a Q-former to compress long videos into a fixed number of tokens \cite{he2024ma}. Similarly, Song et al. use a sliding window approach to achieve long-short-term memory compression \cite{song2024moviechat}. Token shuffling is another method proposed to compress long video tokens in VideoChat-T \cite{zeng2024timesuite}. To capture fine-grained information, VTimeLLM \cite{huang2024vtimellm} introduces a boundary-aware three-stage training strategy, while TimeSuite \cite{zeng2024timesuite} uses temporal adaptive position encoding. VideoLLaMA 2 \cite{cheng2024videollama} incorporates a tailor-made spatial-temporal convolution connector, and Slowfast-LLava \cite{xu2024slowfast} designs a low-speed feature sampler to enhance temporal awareness in visual representation. Despite these advancements, these methods still face challenges such as information loss and increased parameters and inference steps. To address these issues, we adopt an LLM-based long-video agent system that avoids directly processing the entire video, leveraging various tools to acquire information at different granularities.
\subsection{LLM-based agent system for video understanding}
LLM-based agent systems have become an important application for large language models \cite{wang2024survey, liu2024toolace, guo2024large}. The computer vision community has begun exploring LLM-based agent approaches in video understanding \cite{gao2023assistgpt, yang2024doraemongpt}. VideoINSTA \cite{liao2024videoinsta} introduces event-based temporal reasoning and content-based spatial reasoning to improve the LLM's reasoning ability for long videos. In \cite{fan2025videoagent}, a structured spatio-temporal information memory is proposed to support the video agent system. VideoAgent \cite{wang2024videoagent} introduces a framework for interactive reasoning and planning, focusing on the ability to process lengthy visual inputs. While these methods primarily focus on the architecture of the agent system, our approach mimics the human process of video understanding by designing a dedicated CoT with a plan-adjust mode to enhance LLM reasoning and decision-making in long-video scenarios.
\subsection{Bootstrapping  CoT in LLM reasoning}
CoT has been widely used to improve reasoning in LLMs \cite{xu2024llava, wei2022chain, zhang2022automatic}. Much of the research focuses on designing reward processes, including tree search using reinforcement learning \cite{zhang2024rest}, optimization of Q-value ranking \cite{li2024process}, training process advantage verifiers \cite{setlur2024rewarding}, and meta-rewarding steps \cite{wu2024meta}. However, these methods often require additional parameters or reasoning steps, making them difficult to implement in video agent systems. In contrast, our approach focuses on the uncertainty in both LLMs and tools, proposing its use to guide the CoT process. This method is straightforward to implement and helps alleviate the hallucinations that commonly occur in agent systems.
\begin{figure*}
    \centering
    \includegraphics[width=1\linewidth]{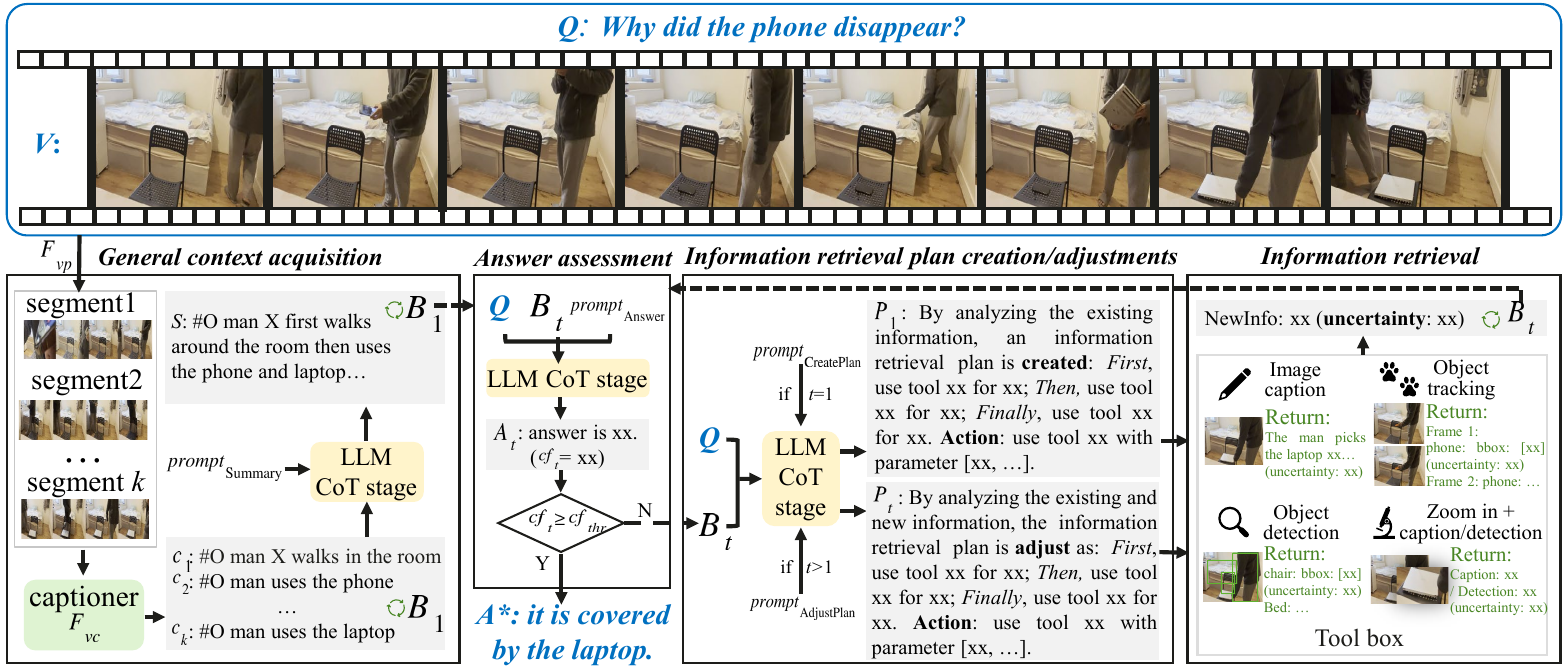}
    \caption{Overview of VideoAgent2. VideoAgent2 answers a question $Q$ about a video $V$ through a pipeline consisting of four phases: general context acquisition, answer assessment, information retrieval plan creation/adjustment, and information retrieval. Details of each phase are introduced in Section \ref{sec:method}.}
    \label{fig:main_fig}
\end{figure*}
\section{Method}\label{sec:method}
In this section, we introduce the proposed VideoAgent2, with the overall framework illustrated in Fig. \ref{fig:main_fig}. Compared to the previous SOTA agent-based approach, VideoAgent \cite{wang2024videoagent}, our method introduces several key enhancements:
(1) Uncertainty-aware CoT reasoning. In VideoAgent, the LLM continuously provides descriptions of the additional information needed based on the general context and the given question. CLIP is then used to retrieve relevant frames by computing similarities, and the captions of these frames serve as new information for reasoning. However, this approach only increases the quantity of retrieved frames and lacks the ability to progressively reason from coarse- to fine-grained levels within a specific scope, as humans naturally do. Moreover, the localization of key information in VideoAgent relies entirely on CLIP, which can introduce noise and bias. In contrast, our method adopts a plan-adjust CoT framework that guides the LLM to analyze and iteratively refine its retrieval strategy, progressively narrowing down from coarse- to fine-grained information. This process incorporates uncertainty estimation when adjusting the retrieval plan and integrates outputs from all tools to produce more reliable answers. (2) New pipeline design. To address the limitations in VideoAgent—where uniform sampling for context retrieval may overlook critical information and cause subsequent retrieval failures—VideoAgent2 adopts a segment-based caption and summarization approach, which will be described in detail in the following section. Furthermore, we  develop a variety of specialized tools tailored for VideoAgent2 to enhance its overall effectiveness.

VideoAgent2 mimics the human process of understanding a long video. Given a long video and a question, a human would first watch the video roughly to gain a general context. If the general context is insufficient for answering the question, they would devise an information retrieval plan based on the question and then explore the relevant segments according to the plan to retrieve enough information to answer the question. Depending on the complexity of the question, this retrieval process may involve multiple steps, each of which is an assessment made to determine the adequacy of the information obtained so far. For clarity, we describe the entire process of VideoAgent2 answering a question $Q$ about a video $V$ through four phases: General context acquisition, Answer assessment, Information retrieval plan creation/adjustments, and Information retrieval. Additionally, we define several key concepts. The answer assessment moment $t$ serves as the state update point for the system. The system's status is represented as ${B_t, P_t, A_t, cf_t \mid 1 \leq t \leq T}$, where $B_t$ is the information memory bank that stores all the information related to $V$, $P_t$ is the plan for retrieving additional information, $A_t$ is the answer generated based on $B_t$, and $cf_t$ is the confidence of the $A_t$.

\subsection{Phase 1: General context acquisition} 
Inspired by human comprehension of long videos, where a person first watches the video roughly to get the general context and then scrutinizes segments that may contain the desired information based on the question, we begin by acquiring the general context information of the target video. There are two main issues to address in this phase: (1) Computational efficiency: The process should not focus on too much detail in the video, as this can significantly affect computational efficiency for long videos and may lead to focusing on irrelevant information without considering the question \cite{he2024ma}. (2) Integrity of information: The general context should contain as much information as possible to avoid failing to identify key segments later. With these considerations in mind, we follow these steps to acquire the general context information:
\begin{itemize}
    \item \textit{Step 1}. The long video, with a length of $L$ seconds, is downsampled to a frame rate of ${fps}_{d}$ and divided into successive segments of $n$ seconds each using a video preprocessor $F_{vp}$.
    \item \textit{Step 2}. Captions for all segments, $C = \{ c_1, c_2, \dots, c_k \}, k = L/n$, are generated by a lightweight video captioner $F_{vc}$. 
    \item \textit{Step 3}: $C$ is input into an LLM to generate a summary $S$ for the entire video. 
\end{itemize}

The issue of computational efficiency (1) is addressed by reducing the total number of frames in \textit{Step 1}. In \textit{Step 2}, instead of uniformly sampling several frames and describing them using an image captioner in VideoAgent, we use a video captioner to preserve the temporal information, preventing the loss of context as mentioned in issue (2). To avoid the captions of each clip becoming too irrelevant or disconnected for the LLM, and inspired by \cite{zhang2024hcqa}, we let the LLM read all the captions and generate a summary of the entire video via CoT prompting. This approach allows the LLM to reflect on the temporal and spatial relationships between all clips, thereby deepening its understanding of the video content. By the end of Phase 1, the general context information $B_1 = \{C, S\}$ is obtained.  Due to the limitations of the video captioner's performance, $B_1$ may lack fine-grained information relevant to the question. Therefore, we first attempt to answer the question directly based on $B_1$ and then evaluate the answer to determine whether additional information retrieval is necessary. We provide a detailed explanation of this process in the next section.

\subsection{Phase 2: Answer assessment}
Humans often assess whether the information they currently have is sufficient to answer a question. We call this process answer assessment. Building on the approach in \cite{wang2024videoagent}, we refine the two-step prediction-evaluation process into a single-step prediction-evaluation process for reducing the number of LLM dialog rounds: we ask the LLM to provide an answer ${A_t}$ to the question $Q$ based on all the information stored in $B_t$, and simultaneously generate a confidence level $cf_t$ for the answer (with a predefined range from 0 to 5). The action for Phase 2 is determined by comparing the answer confidence $cf_t$ to a manually set threshold ${cf}_{thr}$:

\begin{itemize} \item \textit{Action 1}: If $cf_t \geq {cf}_{thr}$, the LLM is considered to have sufficient information to answer the question, and it will confirm the final answer $A^*$ as $A_t$. \item \textit{Action 2}: If $cf_t < {cf}_{thr}$, the LLM is deemed to have insufficient information, and additional information retrieval is required. This leads to Phase 3 for information retrieval plan creation/adjustments. \end{itemize}

In this phase, we leverage the uncertainty in the LLM (referred to as self-reflection) to determine whether the current information is adequate \cite{shinn2024reflexion}.  In cases of insufficient information, we guide the LLM to create or adjust the information retrieval plan, enabling it to build a chain of thought that progressively plans tool calls and interprets the retrieved information.
\subsection{Phase 3: Information retrieval plan creation/adjustments}
When humans determine that the current information is insufficient to answer a question, they typically locate key ranges based on recall and then scrutinize the content to find the desired information. It is essential to note that humans first think about how to retrieve the necessary information (i.e., creating an information retrieval plan based on the question) before examining the relevant segments. For example, with the question "What is the animal in the photo that the man is holding in his hand at the beginning of the video?", a human might devise the following plan:

\begin{itemize} 
\item Step 1: Find the man in the 0-10s. 
\item Step 2: Locate the man's right hand. 
\item Step 3: Examine the photo in the right hand closely. 
\end{itemize}

In practice, the user would adjust the plan as new information is obtained at each step. For instance, if no men are found in the 0-10s segment, Step 2 would be modified to "find men in 10s-20s". Or, if the photo in Step 3 is blurry, Step 4 might be added: "Locate the photo in the next frame and examine it closely". Inspired by this process, we propose a plan-adjust CoT mode, where the LLM first creates an information retrieval plan based on general context and the question, then continuously adjusts the plan based on new information acquired at each step to retrieve the desired information. The actions in Phase 3 are as follows:
\begin{itemize} 
\item \textit{Action 1}: If $t=1$, the system is in the initial state. The LLM is asked to create an information retrieval plan $P_1$ based on $B_1$ and $Q$. 
\item \textit{Action 2}: If $t>1$, the system is in the retrieval process. The LLM is asked to adjust the plan $P_{t-1}$ and generate a new plan $P_t$ based on $B_t$ and $Q$.
\end{itemize}

Unlike the approach in VideoAgent \cite{wang2024videoagent}, which employs a CLIP model to determine the retrieval range by computing similarities between textual descriptions and image features—resulting in a significant additional computational burden—we let the LLM directly decide two key parameters for each tool invocation (for system stability, only one tool is invoked per step): (1) the target frame range for information retrieval, which the LLM can determine based on $C$, and (2) the specific parameters for the tool itself. This design enables our method to flexibly combine tools with different levels of granularity, allowing for progressive localization of key information to achieve more accurate retrieval.
\subsection{Phase 4: Information retrieval}
In this phase, the LLM calls the tool to retrieve new information according to the created or adjusted plan $P_t$. It is crucial to note that the tools themselves may be prone to errors or noise. For example, object detection tools may misidentify rare targets, and image captioning tools may inaccurately describe events in images. These errors affect the LLM's adjustments to the information retrieval plan, leading to increasing errors as the process continues. To address this, we propose using uncertainty to guide the CoT process. Specifically, we let each tool generate a confidence score when it returns the retrieved information. The methodology for obtaining confidence scores for different tools is outlined in Section \ref{exp_setting}. This design enables the LLM to take into account both the content and the reliability of the retrieved information when refining the retrieval plan and integrating all information to make the final decision. Combined with $cf_t$, a complete uncertainty-aware CoT process is established in VideoAgent2, where both the LLM's and the tool's uncertainties are factored into information acquisition and analysis, enhancing the reliability of the entire system.

The VideoAgent2 process is summarized in Algorithm 1.

\begin{algorithm}
\caption{VideoAgent2}
\label{alg:videoagent}
\textbf{Require:} long video $V$, question $Q$, LLM $F_{llm}$, video preprocessor $F_{vp}$, video captioner $F_{vc}$, video tools $F_{vt1}, F_{vt2}, \cdots, F_{vtM}$, max number of answer assessments $T$, confidence threshold $cf_{thr}$, \\
\textbf{Ensure:} information memory bank, information retrieval plan and predicted answer $\{B_t, P_t, A_t|1 \leq t \leq T\}$
\begin{algorithmic}[1]
\STATE $C = \{c_1,c_2,\dots ,c_k\} \leftarrow F_{vc}(F_{vp}(V))$
\STATE $S \leftarrow F_{llm}(C, prompt_{\textbf{Summary}}), B_1 = \{C, S\}$
\FOR{$t = 1$ \TO $T$}
    \STATE $A_t, cf_t \leftarrow F_{llm}(B_t, Q, prompt_{\textbf{Answer}})$
    \IF{$cf_t\geq {cf}_{thr}$}
        \STATE \textbf{break}
    \ELSE
        \IF{$t==1$}
            \STATE $P_t \leftarrow F_{llm}(B_t, Q, prompt_{\textbf{CreatePlan}})$ 
        \ELSE 
            \STATE $P_t \leftarrow F_{llm}(B_t, Q, P_{t-1},prompt_{\textbf{AdjustPlan}})$
        \ENDIF 
        \STATE $\text{NewInfo}  \leftarrow F_{vtm}(P_{t})$
        \STATE $B_{t+1} \leftarrow \text{Merge}(B_t, \text{NewInfo})$
    \ENDIF
\ENDFOR
\STATE \textbf{return} $A^* = A_t$
\end{algorithmic}
\end{algorithm}
\section{Experiments}
We first introduce the experimental settings and then present the experimental results of our methods and baselines, demonstrating the effectiveness of our method.
\subsection{Experimental Setting}\label{exp_setting}
\noindent{\textbf{Datasets.}} We follow the strong baselines \cite{wang2024videoagent, zhang2023simple} using three well-established datasets to evaluate the proposed method. These datasets are described as follows:
\begin{itemize}
    \item Egoschema \cite{mangalam2023egoschema}. EgoSchema consists of over 5,000 human-curated multiple-choice question-answer pairs, spanning more than 250 hours of real video data, and covering a broad range of natural human activities and behaviors. The dataset includes a test set with a subset of 500 questions having publicly available labels. The full set of questions is evaluated exclusively on the official leaderboard. We compare VideoAgent2 with top methods in published papers on the public leaderboard, following the evaluation protocol in \cite{zhang2024hcqa}.
    \item  NExT-QA \cite{xiao2021next}. The NExT-QA dataset includes 5,440 natural videos with 4,880 multiple-choice questions of three types: action reasoning, temporal action reasoning, and common scene comprehension. The validation set of NExT-QA contains 570 videos and 5,000 multiple-choice questions. A challenging subset of NExT-QA, the ATP-hard subset, includes the hardest QA pairs requiring long-term temporal reasoning. We primarily compare VideoAgent2 with top zero-shot methods on the NExT-QA validation set and ATP-hard subset, following \cite{wang2024videoagent, buch2022revisiting}, and also report the top results of supervised methods. 
    \item IntentQA \cite{li2023intentqa}.The IntentQA dataset contains 4,303 videos and 16,000 multiple-choice question-answer pairs, focusing on reasoning about people's intent in videos. The IntentQA test set consists of 567 videos with 2,134 questions. We primarily compare VideoAgent2 with top zero-shot methods on the IntentQA test set, following the methodology in \cite{zhang2023simple}, and also present the top results of supervised methods. 
\end{itemize}
\noindent{\textbf{Metric.}} Since the tasks in all datasets are multiple-choice, accuracy is used as the evaluation metric by following \cite{zhang2023simple, wang2024videoagent}, defined as the number of correct answers divided by the total number of questions.

\noindent{\textbf{System parameters.}} $fps_d$ is set to 1, and $n$ is set to 4 to ensure computational efficiency. $cf_{thr}$ is set to 5 to ensure sufficient confidence in the final answer. $T$ is set to 5 to prevent dead ends in extreme situations.

\noindent{\textbf{LLM and tools selection.}} We select GPT-4o \cite{hurst2024gpt} as the LLM $F_{llm}$, following \cite{zhang2024hcqa}. For video captioning, we use the lightweight captioner LaViLa \cite{zhao2023learning}, as suggested in \cite{wang2024videoagent, zhang2024hcqa}. Additionally, we design or select the following tools and describe how to obtain the confidence score for each tool:
\begin{itemize}
    \item  Image caption. We use GPT-4o \cite{hurst2024gpt} to generate captions for specified frame images. We prompt GPT-4o  to automatically generate a confidence score ranging from 0 to 1 after each sentence or clause in the caption (e.g., "The image shows a person sewing fabric (confidence=0.9), using a needle and thread (confidence=0.95). A green cutting mat is visible on the table (confidence=0.85)."). 
    \item  Object detection. We design an object detection tool based on SAM2 \cite{ravi2024sam2} and Yolov11 \cite{yolo11_ultralytics} for specified frame images. We use the confidence score provided by SAM2 for each detected object. 
    \item  Image zoom in + Image caption. We use OpenCV \cite{culjak2012brief} to zoom in on a specified area of an image frame and generate a caption for the zoomed-in image. We provide the confidence score for the zoomed-in image using the same method as the image caption tool. 
    \item Image zoom in + Object detection. We use OpenCV to zoom in on a specified area of an image frame and perform object detection on the zoomed-in image. The confidence score is obtained using the same method as the object detection tool. 
    \item  Object tracking. We design an object tracking tool based on SAM2 \cite{ravi2024sam2} and Yolov11 \cite{yolo11_ultralytics} to track a specified object within a given frame range. The confidence score is derived from the confidence provided by SAM2 in each frame.
\end{itemize}
Detailed parameters/descriptions of these tools are provided in the Appendix. 

\noindent{\textbf{Baselines.}} We compare VideoAgent2 with work that performed strongly on each dataset by following \cite{wang2024videoagent, zhang2023simple}. 
\subsection{Main results}
VideoAgent2 achieves the SOTA  among zero-shot methods on all datasets (including subsets), which are shown in Tables \ref{tab:exp_mainresult_egoschema}, \ref{tab:exp_mainresult_nextqa} and \ref{tab:exp_mainresult_intentqa}, respectively. 
\begin{table}[h]
    \centering
    \caption{Results on EgoSchema dataset}
    \begin{tabular}{lccc}
        \toprule
       Method & full test set & sub test set \\
        \midrule
        LongViViT \cite{papalampidi2024simple}  &33.3 & 56.8\\
        LLoVi \cite{zhang2023simple}               & 50.3 &57.6 \\
        VideoAgent \cite{wang2024videoagent}     & 54.1 &60.2 \\
        GPT-4V \cite{balavzevic2024memory} &  55.6 &63.5\\
        
        ProViQ \cite{choudhury2023zero}             & 57.1 & 61.2 \\
        InternVideo2 \cite{wang2024internvideo2}  & 60.2 & - \\
        Gemini 1.5 Pro \cite{team2024gemini}    & 63.2 & - \\
        LifelongMemory \cite{wang2023lifelongmemory}   & 64.7  &72.0 \\
        iLearn \cite{zhang2024hcqa}     &  74.6 & 58.8\\
       \textbf{VideoAgent2 (ours)} &  \textbf{75.4} &  \textbf{80.6} \\
        \bottomrule
    \end{tabular}
    \label{tab:exp_mainresult_egoschema}
\end{table}

\begin{table}[h]
    \centering
    \caption{Results on NExT-QA dataset}
    \begin{tabular}{lccc}
        \toprule
       Method & \tabincell{c}{val set} & \tabincell{c}{ATP-hard \\ subset} \\
        \midrule
         \textit{Supervised} \\
        ViLA \cite{wang2024vila}        &74.4   & -     \\
        VideoChat2 \cite{li2024mvbench}  &79.5  & 68.2 \\
        LLaVA-OV \cite{li2024llava}     &  80.2 & -\\
        LinVT-Qwen2-VL \cite{gao2024linvt} &  85.5 & 69.1 \\
        \midrule
        \textit{Zero-shot} \\
        ViperGPT \cite{suris2023vipergpt}  & 60.0 &- \\
        SeViLA \cite{yu2024self}      & 63.6 & 50.8 \\
        VideoAgent \cite{wang2024videoagent}   & 71.3  & 58.4 \\
         LLoVi \cite{zhang2023simple}    &  73.8 & - \\
       \textbf{VideoAgent2 (ours)} &  \textbf{80.5} &  \textbf{68.2} \\
        \bottomrule
    \end{tabular}
    \label{tab:exp_mainresult_nextqa}
\end{table}

\begin{table}[h]
    \centering
    \caption{Results on IntentQA dataset}
    \begin{tabular}{lcc}
        \toprule
       Method & test set \\
        \midrule
         \textit{Supervised} \\   
        IntentQA \cite{li2023intentqa}     &  57.6 \\
        Human \cite{li2023intentqa} &  78.5 \\
         VideoChat2 \cite{li2024mvbench}  &81.9 \\
        \midrule
        \textit{Zero-shot} \\
        LLoVi \cite{zhang2023simple}  & 67.1  \\
        VideoAgent \cite{wang2024videoagent}      & 69.3  \\
        LVNet \cite{park2024too}   & 71.1  \\
         ENTER \cite{ayyubi2025enter}    &  71.5\\
       \textbf{VideoAgent2 (ours)} &  \textbf{73.9}  \\
        \bottomrule
    \end{tabular}
    \label{tab:exp_mainresult_intentqa}
\end{table}

As shown in Tables \ref{tab:exp_mainresult_egoschema}, \ref{tab:exp_mainresult_nextqa}, and \ref{tab:exp_mainresult_intentqa}, VideoAgent2 achieves the best results across all methods for the EgoSchema dataset, and for zero-shot methods on the NExT-QA and IntentQA datasets. Compared to previous SOTA zero-shot results, the accuracy improves by 0.8$\%$ and 8.6$\%$ for the full test set and the sub test set of the EgoSchema dataset, 6.7$\%$ and 9.8$\%$ for the NExT-QA validation set and ATP-hard subset, and 2.4$\%$ for the IntentQA test set. It is worth noting that iLearn performs very well on the EgoSchema full test set but poorly on the subset. We believe this is due to the method's complete reliance on the output of the video captioner, which is not robust to the changes in the data distribution. In contrast, VideoAgent2 shows outstanding performance on both sets, highlighting the robustness of the method. On the NExT-QA dataset, we observe that VideoAgent2 significantly improves the SOTA results for zero-shot methods and even approaches the SOTA results in supervised methods on the ATP-hard subset. This demonstrates that the model possesses strong capabilities in causal, temporal, and descriptive problem analysis. Additionally, VideoAgent2 performs very well among all zero-shot methods on the IntentQA dataset, approaching human-level performance.

\subsection{Case study} 
\begin{figure*}
    \centering
    \includegraphics[width=0.9\linewidth]{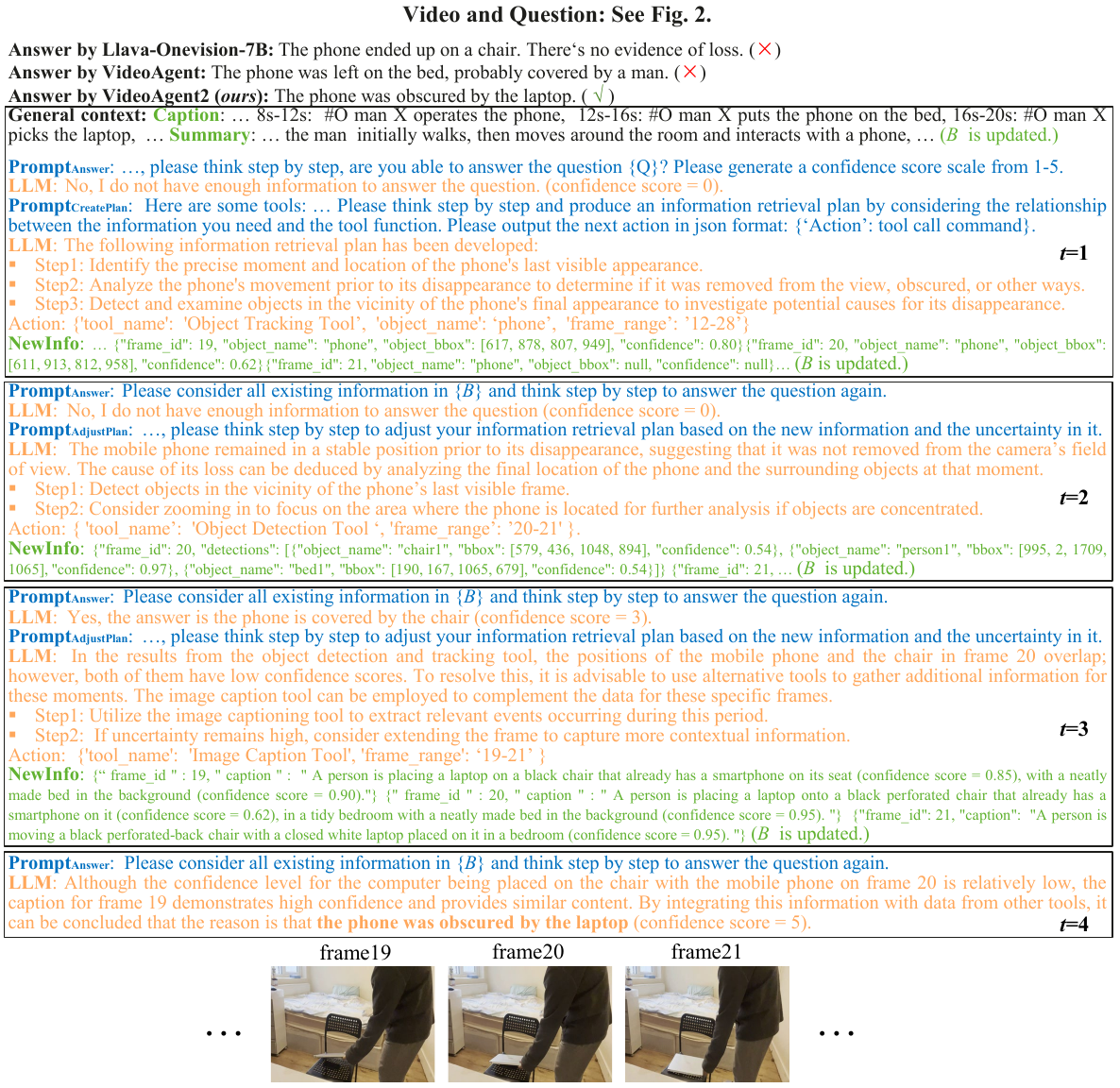}
    \caption{Case study of VideoAgent2. The video and associated question are presented in Fig. \ref{fig:main_fig}. Both the popular MLLM Llava-OneVision and the SOTA agent baseline, VideoAgent, fail to provide the correct answer. In contrast, our proposed VideoAgent2 correctly answers the question through three tool calls and four answer assessments. VideoAgent2 leverages the information and uncertainty  provided by the tools, enabling the LLM to continuously adjust its information retrieval plan, and make
more reliable decisions when synthesizing the final answer.}
    \label{fig:case_study}
\end{figure*}

We use the video and question in Fig. \ref{fig:main_fig} as a case study. Fig. \ref{fig:case_study} illustrates how VideoAgent2 answers this question, providing a clear example of the proposed approach. Due to space limitations, some information is omitted using $...$, and more details are shown in the Appendix. Based on Fig. \ref{fig:case_study}, we make the following observations:
\begin{itemize} 
\item VideoAgent2 answers the question through three tool calls, four answer assessments and primarily focuses on three frames, demonstrating high frame efficiency. In comparison, VideoAgent \cite{wang2024videoagent} uses information from 15 frames and still fails to provide the correct answer.  The MLLM Llava-Onevision-7B also gives the wrong answer.
\item The proposed plan-adjust CoT excels in handling this complex problem. From $t=1$ to $t=3$, rather than following a fixed retrieval plan, VideoAgent2 adjusts the information retrieval scheme based on newly acquired data, progressing from coarse-grained to fine-grained information acquisition. 
\item The uncertainty-guided CoT process effectively addresses the noise introduced by the tools and enables seamless integration of different tools within the agent system to yield more reliable answers. A notable issue is that Caption 12s-16s in the general context incorrectly describes the man’s action as placing the mobile phone on the bed. This mistake is avoided by implementing the retrieval plan. In NewInfo (t=2), the object detection model fails to detect the phone and the laptop, leading the LLM to incorrectly interpret that the phone is covered by the chair. However, the uncertainty in both the LLM’s answer and the tool’s return value successfully guides the LLM to adjust the retrieval plan and incorporate new tools. Similarly, when integrating information from different tools to make the final judgment at $t=4$, the LLM correctly compares the content of frame19 and frame20, as well as their associated confidence scores, to give an accurate final answer. In summary, the proposed uncertainty-aware CoT effectively mitigates noise and
hallucination in both the LLM and tools, allowing the LLM to refine its information retrieval strategy and make more reliable decisions when synthesizing final answers. 
\end{itemize}

\subsection{Analysis of tool call}
\begin{figure}
    \centering
    \includegraphics[width=0.8\linewidth]{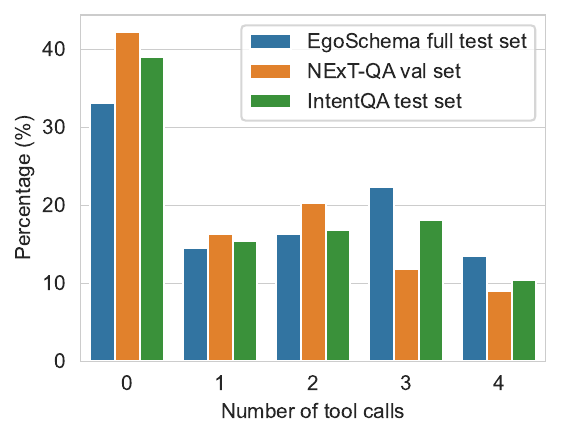}
    \caption{Proportion of samples with different number of tool calls in different datasets. A tool call number of 0 means that for this sample, VideoAgent2 has obtained enough information from the general context information $B_1$ to answer the question without the need for new information retrieval. The maximum tool call number is equal to $T-1$, $T$ is set to 5 in our experiment.}
    \label{fig:average_number_tool_call}
\end{figure}

The key feature of VideoAgent2 is its uncertainty-aware CoT process, which invokes tools through the plan-adjust mode. To better understand this process, we analyze tool calls across different datasets and question types.  \\
\noindent{\textbf{Number of tool calls for different datasets.}} 
We count the number of tool calls for each sample in the EgoSchema full test set, NExT-QA validation set, and IntentQA test set, and show the ratio of each count in Fig. \ref{fig:average_number_tool_call}. A tool call count of 0 means that VideoAgent2 has gathered enough information from the general context information $B_1$ to answer the question without requiring new information retrieval. The maximum number of tool calls is equal to $T-1$ (where $T$ is set to 5 in our experiment). From Fig. \ref{fig:average_number_tool_call}, we observe that the number of samples with zero tool calls in all datasets accounts for the highest percentage, with all datasets exceeding 30$\%$. We attribute this to the fact that video clip captions ensure the completeness of the context information, and the model also deepens its understanding of the spatio-temporal content of the full video by summarizing clip captions. Additionally, we find that the proportion of samples with zero tool calls in the EgoSchema full test set is the lowest among the three datasets, with a significantly larger proportion of samples having 3 or 4 tool calls compared to the other datasets. We believe this is due to the average length of the videos in this dataset being much longer than those in the other datasets, which also results in the higher certificate length and difficulty \cite{mangalam2023egoschema}. For the NExT-QA and IntentQA datasets, the highest non-zero tool call numbers are 2 and 3, respectively. We attribute this to the fact that the IntentQA dataset is constructed using the more challenging parts of the NExT-QA dataset.

\begin{figure}
    \centering
    \includegraphics[width=0.9\linewidth]{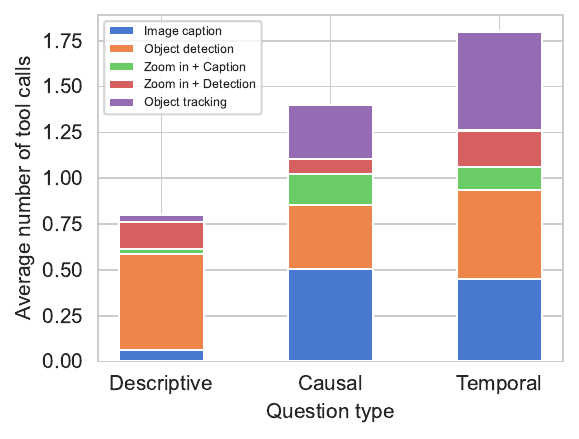}
    \caption{The average number of tool calls  for each  type of question and the average number of calls for each tool in NExT-QA val set.}
    \label{fig:tool_call_question_types}
\end{figure}

\noindent{\textbf{Number of tool calls for different types of questions.}} The NExT-QA dataset categorizes questions into three types: (1) \textit{Causal Questions}, which explore cause-effect relationships in a video, asking why an action occurs or how it leads to an outcome, with both the cause and effect being observable; (2) \textit{Temporal Questions}, which assess the sequence of actions, asking what happened before, during, or after an event, and emphasize multi-object interactions rather than single-object timelines; and (3) \textit{Descriptive Questions}, which focus on scene elements such as locations, objects, and key actions. We calculated the average number of tool calls per sample for each question type, as well as the average number of calls for each tool in the NExT-QA validation set. The results are shown in Fig. \ref{fig:tool_call_question_types}. For descriptive questions, the minimum average number of tool calls is only 0.8, with the vast majority of calls directed to object detection. This is because descriptive questions are primarily concerned with factual descriptions, which are well generalized in the general context information, requiring only object detection tools for basic image frame information retrieval when necessary. In contrast, temporal questions involve the most tool calls. The distribution of tools invoked is similar to that for causal questions, but temporal questions feature significantly more invocations of the object tracking model. We attribute this to the fact that temporal questions require more complex reasoning about spatio-temporal information. Additionally, we observe a higher usage of the image caption tool for both causal and temporal questions. This is likely due to the tool's ability to visualize the details of events in image frames, which is crucial for questions requiring event reasoning, aligning with findings in \cite{wang2024videoagent}.

\subsection{Ablation study}
We perform comprehensive ablation experiments to demonstrate the effectiveness of the proposed approach.


\noindent{\textbf{\noindent{\textbf{Ablation of uncertainty-aware CoT.}}}}  We compare four experimental settings to analyze the role of the uncertainty-aware CoT in VideoAgent2: (1) disabling tools from generating confidence scores, (2) disabling plan adjustment, i.e., the LLM develops an information retrieval plan and executes it without any modifications, (3) disabling the entire CoT system, i.e., the LLM directly calls tools and generates answers without the CoT process, and (4) disabling all tool calls, i.e., the LLM answers the question solely based on $B_1$. Note that each setting is built incrementally upon the previous one. We perform this experiment on the EgoSchema full test set, NExT-QA validation set, and IntentQA test set, and the results are presented in Table \ref{tab:ablation_COT}. From Table \ref{tab:ablation_COT}, we observe that setting (1) and (2) have the largest negative impact on the performance of VideoAgent2, with accuracy decreasing by 4.3$\%$ and 4.0$\%$, respectively. We attribute this outcome to the importance of uncertainty quantification of tools and the iterative adjustment of information retrieval plans within the uncertainty-aware CoT process. These components enable the LLM to account for the reliability of the retrieved information during reasoning, dynamically adjust its retrieval plans, and comprehensively consider all available information to produce more accurate answers. The performance degradation observed in setting (3) and (4) further highlights the importance of tool call and effective call planning in VideoAgent2.
\begin{table}[h]
    \centering
    \caption{Results of VideoAgent2 under different settings}
     \resizebox{1\linewidth}{!}{
    \begin{tabular}{lcccc}
        \toprule
       \textbf{Method} & \tabincell{c}{EgoSchema\\ fullset} &  \tabincell{c}{NExT-QA \\ val set}   &\tabincell{c}{IntentQA \\  test set} &average\\
        \midrule
                      VideoAgent2 &75.4  &80.5 & 73.9 & 76.6 \\
        \tabincell{c}{setting 1}  &71.3& 76.0&	69.5 & 72.3 \\
        \tabincell{c}{setting 2}  &67.1	&71.5&	66.3 & 68.3  \\
        \tabincell{c}{setting 3}  &65.6	&69.8&	65.6 & 67.0 \\
        \tabincell{c}{setting 4}  &60.2& 68.9&63.1& 64.1\\
        \bottomrule
    \end{tabular}}
    \label{tab:ablation_COT}
\end{table}

\noindent{\textbf{\noindent{\textbf{Ablation of different LLM}}}} We explore the impact of using different LLMs within VideoAgent2. In Table \ref{tab:ablation_LLM}, we compare the performance of VideoAgent2 on the EgoSchema full test set, NExT-QA validation set, and IntentQA test set using popular commercial LLMs such as GPT-4o \cite{hurst2024gpt}, GPT-4 \cite{achiam2023gpt}, and open-source LLMs Deepseek-V3 \cite{liu2024deepseek} and Llama3.3-70B \cite{dubey2024llama}. Our findings indicate that GPT-4o demonstrates superior reasoning capability compared to the other models, achieving the best overall performance. Notably, the open-source model Deepseek-V3 achieves performance close to GPT-4o, offering users more flexibility in selecting LLMs.
\begin{table}[h]
    \centering
    \caption{Results of VideoAgent2 with different LLM}
    \begin{tabular}{lccc}
        \toprule
       LLM &\tabincell{c}{EgoSchema\\ full test set} &  \tabincell{c}{NExT-QA \\ val set}   &\tabincell{c}{IntentQA \\  test set}\\
        \midrule
        GPT-4o&75.4  &80.5 & 73.9 \\
        GPT-4  &74.1  &78.2 & 72.6 \\
        Deepseek-V3& 74.7  & 79.5 & 73.4 \\
        Llama3.3-70B &70.5  &75.3 & 68.6 \\
        \bottomrule
    \end{tabular}
    \label{tab:ablation_LLM}
\end{table}

\section{Conclusion}
In this paper, we introduce VideoAgent2, an enhanced LLM-based agent system designed for effective long-form video understanding through a novel uncertainty-aware Chain-of-Thought (CoT) mechanism. VideoAgent2 addresses critical challenges faced by existing video agent systems, including limited reasoning capabilities and susceptibility to errors introduced by external tools. The proposed uncertainty-aware CoT mechanism enables adaptive and robust reasoning by incrementally refining information retrieval plans, guided by the uncertainty estimation derived from both internal assessments by the LLM and external tool outputs. Extensive experiments conducted on prominent benchmarks—including EgoSchema, NExT-QA, and IntentQA—demonstrate that VideoAgent2 achieves state-of-the-art performance, significantly outperforming existing zero-shot methods. Future work will focus on further refining uncertainty estimation methods and exploring additional multimodal integration strategies to continuously improve the generalization and efficiency of LLM-based video understanding systems.

{
    \small
    \bibliographystyle{ieeenat_fullname}
    \bibliography{main}
}
\clearpage
\onecolumn
\setcounter{page}{1}
\setcounter{section}{0}
{\centering
        \Large
        \textbf{\textit{Appendix for} VideoAgent2: Enhancing the LLM-based agent system for
long-form video understanding by uncertainty-aware CoT} \\
}
\renewcommand{\thesection}
{\Alph{section}} 

This document provides more details of our approach, organized as follows:
\begin{itemize}
    \item \S\ A Details of used/designed tools in VideoAgent2
    \item \S\ B Prompts for VideoAgent2
\end{itemize}
\section{Details of used/designed tools in VideoAgent2}
We show more details of used/designed tools in VideoAgent2.
\subsection{Image caption} 
We use GPT-4o  to generate captions for specified frame images. We ask GPT-4o  to automatically generate a confidence score ranging from 0 to 1 after each sentence or clause in the caption by the prompt:
\begin{verbatim}
"You are an assistant that generates descriptive captions for images.
For each sentence or clause in the caption, include a confidence score
in the format (confidence=0.xx) after the description. 
This confidence is from 0 to 1, reflecting your confidence of the caption.
Here is an example:
'The image shows a small kitchen counter with a kettle (confidence=0.94), 
a round black electronic device (confidence=0.85), a loaf of bread (confidence=0.73), 
and some cleaning supplies (confidence=0.95). There is a trash can on the floor 
(confidence=0.85) and a blue tiled backsplash (confidence=0.62).'"
\end{verbatim}
\begin{figure}[h]
    \centering
    \includegraphics[width=0.6\linewidth]{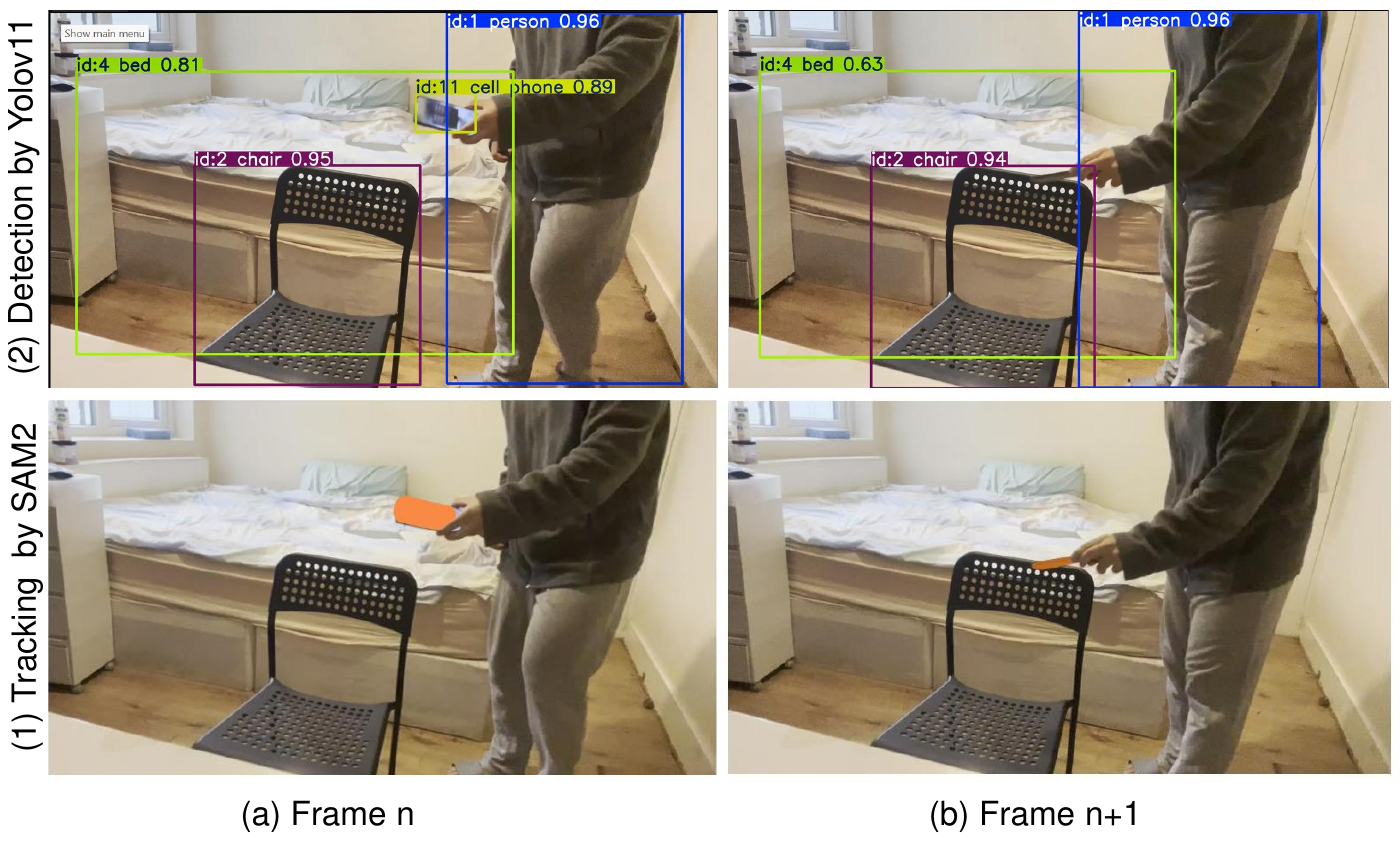}
    \caption{An example of miss detection in object detection result. In (a)(2), the target detection model correctly identifies the mobile phone with high confidence. However, in (b)(2), due to a change in the phone's orientation, the model fails to detect it. By contrast, leveraging SAM2 to track the mobile phone across these two frames effectively addresses this issue, as demonstrated in (a)(1) and (b)(1).}
    \label{fig:appendixA}
\end{figure}
\subsection{Object detection} 
Object detection models such as the Yolo series offer strong performance and speed but still encounter challenges such as missed detections, as illustrated in Fig. \ref{fig:appendixA}. To address this issue, we incorporate SAM2 to construct a multi-round target detection framework that significantly reduces target loss by tracking the specified target across frames. Taking the object detection for the $m$-th frame as an example, the overall process is summarized as follows:

\begin{enumerate} 
\item The frame range is extended to [$m-\alpha$, $m+\alpha$], and all frames within this range are processed using Yolov11. To maintain computational efficiency, $\alpha$ is set to a small value, typically 5. 
\item For each target, the frame with the highest confidence score provided by Yolov11, exceeding a predefined threshold, is selected. The bounding box from this frame is then used to initialize SAM2, which performs bi-directional tracking to retrieve the target's information in the $m$-th frame. We calculate the bounding box as the final detection result based on the mask provided by SAM2 in the $m$-th frame.
\end{enumerate}
\subsection{Object tracking} 
SAM2 demonstrates strong performance in target tracking. However, it requires manual specification of the target’s initial position before starting. To enable object tracking in VideoAgent2 by simply specifying the name of the item, we use Yolov11 to automatically initialize SAM2. Taking the task of tracking the target “mobile phone” within the frame range [$m$, $n$] as an example,  the overall process is as follows:
\begin{enumerate}
\item Apply Yolov11 to detect the target “mobile phone” in frames [$m$, $n$].
\item If the detection confidence exceeds a predefined threshold, use the detected bounding box to initialize SAM2. Bi-directional tracking is then performed to obtain the complete trajectory of the target across the frame range [$m$, $n$].
\end{enumerate}

\section{Details of prompt in VideoAgent2}
We show all the prompts in VideoAgent2.
\begin{itemize}
    \item  $prompt_\text{Summary}$
    \begin{verbatim}
"A long video is segmented into consecutive 4-second clips.
Given the captions for all the clips, {C}, please summarize them into a coherent 
description of the entire video.
Please approach this task step by step, carefully considering the temporal and 
spatial relationships between the  content in each clip during the summarization
process.
Note for captions:
- '#C'  indicates actions performed by the camera wearer 
(the person who recorded the video while wearing the camera).
- '#O'  indicates actions performed by someone other than the camera wearer."
    \end{verbatim}
    \item  $prompt_\text{Answer}$
    \begin{verbatim}
"Now we have a memory bank {B} which stores all the information of a video.
In this bank:
- 'Caption' is the caption of all consecutive sub-segments of the video.
- 'Summary' is the summarized description of the video.
- 'Tools return value' is the retrieved information by some tools.

Please think step by step. Are you able to answer the question {Q}?If you don't 
think  there is enough information to answer the question,  please reply as 'No, 
I do not  have enough information to answer the question. (confidence score = 0)'. 
If you can  answer the question, please reply as 'Yes, the  answer is xx, 
(confidence  score = xx)',  note that you need to generate a confidence  score for
your answer,  scaled from 1-5."
    \end{verbatim}
    
    \item  $prompt_\text{CreatePlan}$
    \begin{verbatim}
"To assist you in answering the question more effectively, I have provided some 
tools.  Below are tool descriptions, notes on using tools, and the call command 
format:

1. Image Caption Tool
   - Function: Generates captions for specific image frames.
   - Usage: Specify a single frame index or a range of frames.
   - Return Values: A list of dictionaries, each containing the frame_id and the 
   caption (e.g., {'frame_id': 'xx', 'caption': 'xx'}). A confidence score is 
   provided for each sentence or clause in the caption.

2. Object Detection Tool
   - Function: Identifies all objects within specific image frames and provides 
   their bounding boxes.
   - Usage: Specify a single frame index or a range of frames.
   - Return Values: A list of dictionaries, each containing the frame_id and the 
   detection results (e.g., {'frame_id': 'xx', 'det_info': {'id': 'xx', 'name': 
   'xx', 'bbox': '[xmin, ymin, xmax, ymax]', 'confidence': 'xx'}}).

3. Image Zoom in and Caption Tool
   - Function: First zoom in on an area of an image frame and then generate a 
   caption.
   - Usage: Specify a single frame index and the bbox of the area you are 
   interested in.
   - Return Values: A list of dictionaries, each containing the frame_id, the bbox 
   of the interested area,  and the caption (e.g., {'frame_id': 'xx', 'bbox': 'xx',
   'caption': 'xx'}). A confidence score is provided for each sentence or clause in 
   the caption.

4. Image Zoom in and Object Detection Tool
   - Function: First zoom in on an area of an image frame and then detect all 
   objects in the area.
   - Usage: Specify a single frame index and the bbox of the area you are 
   interested in.
   - Return Values: A list of dictionaries, each containing the frame_id, the bbox 
   of the interested area,  and the detection results (e.g., {'frame_id': 'xx',
   'bbox': 'xx', 'det_info': {'id': 'xx', 'name': 'xx', 'bbox': '[xmin, ymin, xmax, 
   ymax]', 'confidence': 'xx'}}).

5. Object Tracking Tool
   - Function: Provides the bounding box (bbox) of an object in each frame of a 
   video clip.
   - Usage: Specify the object name and the frame range.
   - Return Values: A list of dictionaries, where each dictionary contains the frame 
   id, object name, bbox and confidence (e.g., {'frame_id': 'xx', 'object_name': 'xx', 
   'bbox': '[xmin, ymin, xmax, ymax]', 'confidence': 'xx'}).

The call command for the Image Caption Tool is:
{'tool_name': 'Image Caption Tool', 'frame_range': 'frame_id' 
# or 'start frame-end frame'}.

The call command for the Object Detection Tool is:
{'tool_name': 'Object Detection Tool', 'frame_range': 'frame_id' 
# or 'start frame-end frame'}.

The call command for the Image Zoom in and Caption Tool is:
{'tool_name': 'Image Zoom in and Caption Tool', 'frame_range': 'frame_id', 
'bbox': '[xmin, ymin, xmax, ymax]'}.

The call command for the Image Zoom in and Object Detection Tool is:
{'tool_name': 'Image Zoom in and Object Detection Tool', 
'frame_range': 'frame_id', 'bbox': '[xmin, ymin, xmax, ymax]'}.

The call command for the Object Tracking Tool is:
{'tool_name': 'Object Tracking Tool', 'object_name': 'xx', 
'frame_range': 'frame_id' # or 'start frame-end frame'}.

You are allowed to call the tool multiple times to retrieve the information you need,
but only one tool can be called at a time. Please think step by step and first make 
an information retrieval plan  to help you gather the useful information. Consider 
the relationship between the information you need and the tool function. Then please 
output  the first action in the following JSON format: {'Action': 'tool call command'}."
    \end{verbatim}
    
    \item  $prompt_\text{AdjustPlan}$
\begin{verbatim}
"Your answer is not confident enough. Please think step by step to adjust your 
information retrieval plan based on the new information and the uncertainty in it and
output  the first action in the following JSON format:  {'Action': 'tool call command'}."
\end{verbatim}

\end{itemize}

\end{document}